\documentclass[unicode=true,breaklinks=true]{article}
\usepackage[latin9]{inputenc}
\usepackage{float}
\usepackage{amsmath}
\usepackage{amsthm}
\usepackage{graphicx}
\usepackage[unicode=true]
 {hyperref}

\makeatletter

\floatstyle{ruled}
\newfloat{algorithm}{tbp}{loa}
\providecommand{\algorithmname}{Algorithm}
\floatname{algorithm}{\protect\algorithmname}

  \theoremstyle{definition}
  \newtheorem{defn}{\protect\definitionname}

\usepackage{style/aixi}
\usepackage{amssymb}
\usepackage{style/ijcai17}
\usepackage{algorithm,algpseudocode}
\usepackage{times}

\usepackage{todonotes}
\usepackage[small]{caption}

\makeatother

  \providecommand{\definitionname}{Definition}

\begin{document}

\title{Universal Reinforcement Learning Algorithms: Survey and Experiments}

\author{\textbf{John Aslanides$^{\dagger}$, Jan Leike$^{\ddagger}$}\thanks{Now at DeepMind.}\textbf{,
Marcus Hutter$^{\dagger}$}\\
$^{\dagger}$Australian National University\\
$^{\ddagger}$Future of Humanity Institute, University of Oxford\\
\{john.aslanides, marcus.hutter\}@anu.edu.au, leike@google.com}
\maketitle
\begin{abstract}
Many state-of-the-art reinforcement learning (RL) algorithms typically
assume that the environment is an ergodic Markov Decision Process
(MDP). In contrast, the field of \emph{universal} reinforcement learning
(URL) is concerned with algorithms that make as few assumptions as
possible about the environment. The universal Bayesian agent AIXI
and a family of related URL algorithms have been developed in this
setting. While numerous theoretical optimality results have been proven
for these agents, there has been no empirical investigation of their
behavior to date. We present a short and accessible survey of these
URL algorithms under a unified notation and framework, along with
results of some experiments that qualitatively illustrate some properties
of the resulting policies, and their relative performance on partially-observable
gridworld environments. We also present an open-source reference implementation
of the algorithms which we hope will facilitate further understanding
of, and experimentation with, these ideas.
\end{abstract}

\section{Introduction}

The standard approach to reinforcement learning typically assumes
that the environment is a fully-observable Markov Decision Process
(MDP) \cite{SB:1998}. Many state-of-the-art applications of reinforcement
learning to large state-action spaces are achieved by parametrizing
the policy with a large neural network, either directly (e.g. with
deep deterministic policy gradients \cite{Silver:2014}) or indirectly
(e.g. deep Q-networks \cite{MKSGAWR:2013DQN}). These approaches have
yielded superhuman performance on numerous domains including most
notably the Atari 2600 video games \cite{MKSRV+:2015deepQ} and the
board game Go \cite{DeepMind:2016Go}. This performance is due in
large part to the scalability of deep neural networks; given sufficient
experience and number of layers, coupled with careful optimization,
a deep network can learn useful abstract features from high-dimensional
input. These algorithms are however restricted in the class of environments
that they can plausibly solve, due to the finite capacity of the network
architecture and the modelling assumptions that are typically made,
e.g. that the optimal policy can be well-approximated by a function
of a fully-observable state vector.

In the setting of universal reinforcement learning, we lift the Markov,
ergodic, and full-observability assumptions, and attempt to derive
algorithms to solve this general class of environments. URL aims to
answer the theoretical question: ``making as few assumptions as possible
about the environment, what constitutes optimal behavior?''. To this
end several Bayesian, history-based algorithms have been proposed
in recent years, central of which is the agent AIXI \cite{Hutter:2005}.
Numerous important open conceptual questions remain \cite{Hutter:2009open},
including the need for a relevant, objective, and general optimality
criterion \cite{LH:2015priors}. As the field of artifical intelligence
research moves inexorably towards AGI, these questions grow in import
and relevance. 

The contribution of this paper is three-fold: we present a survey
of these URL algorithms, and unify their presentation under a consistent
vocabulary. We illuminate these agents with an empirical investigation
into their behavior and properties. Apart from the MC-AIXI-CTW implementation
\cite{VNHUS:2011} this is the only non-trivial set of experiments
relating to AIXI, and is the only set of experiments relating to its
variants; hitherto only their asymptotic properties have been studied
theoretically. Our third contribution is to present a portable and
extensible open-source software framework\footnote{The framework is named $\textsc{AIXIjs}$; the source code can be
found at \href{http://github.com/aslanides/aixijs}{http://github.com/aslanides/aixijs}.} for experimenting with, and demonstrating, URL algorithms. We also
discuss several tricks and approximations that are required to get
URL implementations working in practice. Our desire is that this framework
will be of use, both for education and research, to the RL and AI
safety communities.\footnote{A more comprehensive introduction and discussion including more experimental
results can be found in the associated thesis at \href{https://arxiv.org/abs/1705.07615}{https://arxiv.org/abs/1705.07615}.}

\section{Literature Survey}

This survey covers history-based Bayesian algorithms; we choose history-based
algorithms, as these are maximally general, and we restrict ourselves
to Bayesian algorithms, as they are generally both principled and
theoretically tractable. The universal Bayesian agent AIXI \cite{Hutter:2005}
is a model of a maximally intelligent agent, and plays a central role
in the sub-field of universal reinforcement learning (URL). Recently,
AIXI has been shown to be flawed in important ways; in general it
doesn\textquoteright t explore enough to be asymptotically optimal
\cite{Orseau:2010}, and it can perform poorly, even asymptotically,
if given a bad prior \cite{LH:2015priors}. Several variants of AIXI
have been proposed to attempt to address these shortfalls: among them
are entropy-seeking \cite{Orseau:2011ksa}, information-seeking \cite{OLH:2013ksa},
Bayes with bursts of exploration \cite{Lattimore:2013}, MDL agents
\cite{Leike:2016}, Thompson sampling \cite{LLOH:2016Thompson}, and
optimism \cite{SH:2015opt}.

It is worth emphasizing that these algorithms are models of rational
behavior in general environments, and are not intended to be efficient
or practical reinforcement learning algorithms. In this section, we
provide a survey of the above algorithms, and of relevant theoretical
results in the universal reinforcement learning literature. 

\subsection{Notation}

As we are discussing POMDPs, we distinguish between (hidden) states
and percepts, and we take into account histories, i.e. sequences of
actions and percepts. States, actions, and percepts use Latin letters,
while environments and policies use Greek letters. We use $\mathbb{R}$
as the reals, and $\mathbb{B}=\left\{ 0,1\right\} $. For sequences
over some alphabet $\mathcal{X}$, $\mathcal{X}^{k}$ is the set of
all sequences$\approx$strings of length $k$ over $\mathcal{X}$.
We typically use the shorthand $x_{1:k}:=x_{1}x_{2}\dots x_{k}$ and
$x_{<k}:=x_{1:k-1}$. Concatenation of two strings $x$ and $y$ is
given by $xy$. We refer to environments and environment models using
the symbol $\nu$, and distinguish the \emph{true }environment with
$\mu$. The symbol $\epsilon$ is used to represent the empty string,
while $\varepsilon$ is used to represent a small positive number.
The symbols $\rightarrow$ and $\rightsquigarrow$ are deterministic
and stochastic mappings, respectively. 

\subsection{The General Reinforcement Learning Problem}

We begin by formulating the agent-environment interaction. The environment
is modelled as a partially observable Markov Decision Process (POMDP).
That is, we can assume without loss of generality that there is some
hidden state $s$ with respect to which the environment's dynamics
are Markovian. Let the state space $\mathcal{S}$ be a compact subset
of a finite-dimensional vector space $\mathbb{R}^{N}$. For simplicity,
assume that the action space $\mathcal{A}$ is finite. By analogy
with a hidden Markov model, we associate with the environment stochastic
dynamics $\mathcal{D}\ :\ \mathcal{S}\times\mathcal{A}\rightsquigarrow\mathcal{S}$.
Because the environment is in general partially observable, we define
a \emph{percept} space $\mathcal{E}$. Percepts are distributed according
to a state-conditional percept distribution $\nu$; as we are largely
concerned with the agent's perspective, we will usually refer to $\nu$
as the environment itself. 

The agent selects actions according to a \emph{policy $\pi\left(\;\cdot\;\lvert\ae_{<t}\right)$},
a conditional distribution over $\mathcal{A}$. The agent-environment
interaction takes the form of a two-player, turn-based game; the agent
samples an action $a_{t}\in\mathcal{A}$ from its \emph{policy} $\pi\left(\;\cdot\;\lvert\ae_{<t}\right)$,
and the environment samples a percept $e_{t}\in\mathcal{E}$ from
$\nu\left(\;\cdot\;\lvert\ae_{<t}a_{t}\right)$. Together, they interact
to produce a \emph{history}: a sequence of action-percept pairs $h_{<t}\equiv\ae_{<t}:=a_{1}e_{1}\dots a_{t-1}e_{t-1}$.
The agent and environment together induce a telescoping distribution
over histories, analogous to the \emph{state-visit} distribution in
RL:
\begin{equation}
\nu^{\pi}\left(\ae_{1:t}\right):=\prod_{k=1}^{t}\pi\left(a_{k}\lvert\ae_{<k}\right)\nu\left(e_{k}\lvert\ae_{<k}a_{k}\right).\label{eq:state-visit}
\end{equation}

In RL, percepts consist of $\left(\text{observation},\text{reward}\right)$
tuples so that $e_{t}=\left(o_{t},r_{t}\right)$. We assume that the
reward signal is real-valued, $r_{t}\in\mathbb{R}$, and make no assumptions
about the structure of the $o_{t}\in\mathcal{O}$. In general, agents
will have some utility function $u$ that typically encodes some preferences
about states of the world. In the partially observable setting, the
agent will have to make inferences from its percepts to world-states.
For this reason, the utility function is a function over finite histories
of the form $u\left(\ae_{1:t}\right)$; for agents with an extrinsic
reward signal, $u\left(\ae_{1:t}\right)=r_{t}$. The agent's objective
is to maximize expected future discounted utility. We assume a general
class of convergent \emph{discount functions}, $\gamma_{k}^{t}\ :\ \mathbb{N}\times\mathbb{N}\to\left[0,1\right]$
with the property $\Gamma_{\gamma}^{t}:=\sum_{k=t}^{\infty}\gamma_{k}^{t}<\infty$.
For this purpose, we introduce the \emph{value} function, which in
this setting pertains to histories rather than states:

\begin{equation}
V_{\nu\gamma}^{\pi u}\left(\ae_{<t}\right):=\mathbb{E}_{\nu}^{\pi}\left[\left.\sum_{k=t}^{\infty}\gamma_{k}^{t}u\left(\ae_{1:k}\right)\right|\ae_{<t}\right].\label{eq:value}
\end{equation}

In words, $V_{\nu\gamma}^{\pi u}$ is the expected discounted future
sum of reward obtained by an agent following policy $\pi$ in environment
$\nu$ under discount function $\gamma$ and utility function $u$.
For conciseness we will often drop the $\gamma$ and/or $\mu$ subscripts
from $V$ when the discount/utility functions are irrelevant or obvious
from context; by default we assume geometric discounting and extrinsic
rewards. The value of an \emph{optimal policy} is given by the \emph{expectimax}
expression

\begin{eqnarray}
V_{\nu\gamma}^{\star u}\left(\ae_{<t}\right) & = & \max_{\pi}V_{\nu\gamma}^{\pi u}\nonumber \\
 & = & \lim_{m\to\infty}\max_{a_{t}\in\mathcal{A}}\sum_{e_{t}\in\mathcal{E}}\cdots\max_{a_{t+m}\in\mathcal{A}}\sum_{e_{t+m}\in\mathcal{E}}\Biggl(\nonumber \\
 &  & \quad\sum_{k=t}^{t+m}\gamma_{k}^{t}u\left(\ae_{1:k}\right)\prod_{j=t}^{k}\nu\left(e_{j}\lvert\ae_{<j}a_{j}\right)\Biggr)\label{eq:expectimax}
\end{eqnarray}

which follows from Eqs. \eqref{eq:state-visit} and \eqref{eq:value}
by jointly maximizing over all future actions and distributing $\max$
over $\sum$. The optimal policy is then simply given by $\pi_{\nu}^{\star}=\arg\max_{\pi}V_{\nu}^{\pi}$;
note that in general the optimal policy may not exist if $u$ is unbounded
from above. We now introduce the only non-trivial and non-subjective
optimality criterion yet known for general environments \cite{LH:2015priors}:
\emph{weak asymptotic optimality}.
\begin{defn}[Weak asymptotic optimality; Lattimore \& Hutter, 2011]
 Let the \emph{environment class }$\mathcal{M}$ be a set of environments.
A policy $\pi$ is\emph{ weakly asymptotically optimal} in $\mathcal{M}$
if $\forall\mu\in\mathcal{M}$, $V_{\mu}^{\pi}\to V_{\mu}^{*}$ in
mean, i.e.

\[
\mu^{\pi}\left(\lim_{n\to\infty}\frac{1}{n}\sum_{t=1}^{n}\left\{ V_{\mu}^{*}\left(\ae_{<t}\right)-V_{\mu}^{\pi}\left(\ae_{<t}\right)\right\} =0\right)=1,
\]

where $\mu^{\pi}$ is the history distribution defined in Equation
\eqref{eq:state-visit}.
\end{defn}
AIXI is not in general asymptotically optimal, but both BayesExp and
Thompson sampling (introduced below) are. Finally, we introduce the
the notion of \emph{effective horizon}, which these algorithms rely
on for their optimality.
\begin{defn}[$\boldsymbol{\varepsilon}$-Effective horizon; Lattimore \& Hutter,
2014]
Given a discount function $\gamma$, the $\varepsilon$\emph{-effective
horizon} is given by
\begin{equation}
H_{\gamma}^{t}\left(\varepsilon\right):=\min\left\{ H\ :\ \frac{\Gamma_{\gamma}^{t+H}}{\Gamma_{\gamma}^{t}}\leq\varepsilon\right\} .\label{eq:effective-horizon}
\end{equation}

In words, $H$ is the horizon that one can truncate one's planning
to while still accounting for a fraction equal to $\left(1-\varepsilon\right)$
of the realizable return under stationary i.i.d. rewards.
\end{defn}

\subsection{Algorithms\label{subsec:algorithms}}

We consider the class of Bayesian URL agents. The agents maintain
a predictive distribution over percepts, that we call a \emph{mixture
model} $\xi$. The agent mixes$\approx$marginalizes over a class
of models$\approx$hypotheses$\approx$environments $\mathcal{M}$.
We consider countable nonparametric model classes $\mathcal{M}$ so
that

\begin{equation}
\xi\left(e\right)=\sum_{\nu\in\mathcal{M}}\overbrace{p\left(e\lvert\nu\right)}^{\nu\left(e\right)}\underbrace{p\left(\nu\right)}_{w_{\nu}},\label{eq:mixture}
\end{equation}

where we have suppressed the conditioning on history $\ae_{<t}a_{t}$
for clarity. We have identified the agent's credence in hypothesis
$\nu$ with \emph{weights} $w_{\nu}$, with $w_{\nu}>0$ and $\sum_{\nu}w_{\nu}\leq1$,
and we write the probability that $\nu$ assigns to percept $e$ as
$\nu\left(e\right)$. We update with Bayes rule, which amounts to
$p\left(\nu\lvert e\right)=\frac{p\left(e\lvert\nu\right)}{p\left(e\right)}p\left(\nu\right)$,
which induces the sequential weight updating scheme $w_{\nu}\leftarrow\frac{\nu\left(e\right)}{\xi\left(e\right)}w_{\nu}$;
see Algorithm \ref{alg:bayes-rl}. We will sometimes use the notation
$w_{\nu\lvert\ae_{<t}}$ to represent the posterior mass on $\nu$
after updating on history $\ae_{<t}\in\left(\mathcal{A}\times\mathcal{E}\right)^{*}$,
and $w\left(\cdot\,\lvert\:\cdot\right)$ when referring explicitly
to a posterior distribution.
\begin{defn}[AI$\xi$; Hutter, 2005]
AI$\xi$ is the policy that is optimal w.r.t. the Bayes mixture $\xi$:
\begin{align}
\pi_{\xi}^{\star} & :=\arg\max_{\pi}V_{\xi}^{\pi}\nonumber \\
 & =\lim_{m\to\infty}\arg\max_{a_{t}\in\mathcal{A}}\sum_{e_{t}\in\mathcal{E}}\cdots\max_{a_{m}\in\mathcal{A}}\sum_{e_{m}\in\mathcal{E}}\Biggl(\nonumber \\
 & \qquad\left.\sum_{k=t}^{m}\gamma_{k}r_{k}\prod_{j=t}^{k}\sum_{\nu\in\mathcal{M}}w_{\nu}\nu\left(e_{j}\lvert\ae_{<j}a_{j}\right)\right).\label{eq:bayes-expetimax}
\end{align}
\end{defn}
Computational tractability aside, a central issue in Bayesian induction
lies in the choice of prior. If computational considerations aren't
an issue, we can choose $\mathcal{M}$ to be as broad as possible:
the class of all lower semi-computable conditional contextual semimeasures
$\mathcal{M}_{\text{CCS}}^{\text{LSC}}$ \cite{LH:2015computability2},
and using the prior $w_{\nu}=2^{-K\left(\nu\right)}$, where $K$
is the Kolmogorov complexity. This is equivalent to using Solomonoff's
universal prior \cite{Solomonoff:1978} over strings $M\left(x\right):=\sum_{q\ :\ U\left(q\right)=x*}2^{-\left|q\right|}$,
and yields the AIXI model. AI$\xi$ is known to not be asymptotically
optimal \cite{Orseau:2010}, and it can be made to perform badly by
bad priors \cite{LH:2015priors}.

\begin{algorithm}
\begin{algorithmic}[1]
\Require{Model class $\mathcal{M}=\lbrace\nu_1,\dots,\nu_K\rbrace$; prior $w\ :\ \mathcal{M}\to(0,1)$; history $\ae_{<t}$.}
\Statex
\Function{Act}{$\pi$}
\State Sample and perform action $a_t\sim \pi(\cdot\lvert \ae_{<t})$
\State Receive $e_t\sim \nu(\;\cdot\;\lvert \ae_{<t}a_t) $
\For{$\nu\in\mathcal{M}$}
\State $w_\nu \leftarrow \frac{\nu\left(e_t\lvert \ae_{<t}a_t\right)}{\xi\left(e_t\lvert \ae_{<t}a_t\right)}w_\nu$
\EndFor
\State $t \leftarrow t + 1$
\EndFunction
\end{algorithmic}

\caption{Bayesian URL agent {[}Hutter, 2005{]}}
\label{alg:bayes-rl}
\end{algorithm}

\textbf{Knowledge-seeking agents (KSA)}. Exploration is one of the
central problems in reinforcement learning \cite{Thrun92c}, and a
principled and comprehensive solution does not yet exist. With few
exceptions, the state-of-the-art has not yet moved past $\varepsilon$-greedy
exploration \cite{BellemareSOS+:2016,HouthooftCCY+:vime,PAED2017,MNEH2017}.
Intrinsically motivating an agent to explore in environments with
sparse reward structure via \emph{knowledge-seeking} is a principled
and general approach. This removes the dependence on extrinsic reward
signals or utility functions, and collapses the exploration-exploitation
trade-off to simply exploration. There are several generic ways in
which to formulate a knowledge-seeking utility function for model-based
Bayesian agents. We present three, due to Orseau \emph{et al.}:
\begin{defn}[Kullback-Leibler KSA; Orseau, 2014]
\label{def:klksa}The\emph{ KL-KSA} is the Bayesian agent whose utility
function is given by the \emph{information gain}
\begin{eqnarray}
u_{\text{KL}}\left(\ae_{1:t}\right) & = & \text{IG}\left(e_{t}\right)\label{eq:utility-kl}\\
 & := & \text{Ent}\left(w\left(\cdot\lvert\ae_{<t}\right)\right)-\text{Ent}\left(w\left(\cdot\lvert\ae_{1:t}\right)\right).
\end{eqnarray}
\end{defn}
Informally, the KL-KSA gets rewarded for reducing the entropy (uncertainty)
in its model. Now, note that the entropy in the Bayesian mixture $\xi$
can be decomposed into contributions from\emph{ uncertainty} in the
agent's beliefs $w_{\nu}$ and\emph{ noise} in the environment $\nu$.
That is, given a mixture $\xi$ and for some percept $e$ such that
$0<\xi\left(e\right)<1$, and suppressing the history $\ae_{<t}a_{t}$,
\[
\xi\left(e\right)=\sum_{\nu\in\mathcal{M}}\overbrace{w_{\nu}}^{\text{uncertainty}}\underbrace{\nu\left(e\right)}_{\text{noise}}.
\]

That is, if $0<w_{\nu}<1$, we say the agent is \emph{uncertain} about
whether hypothesis $\nu$ is true (assuming there is exactly one $\mu\in\mathcal{M}$
that is the truth). On the other hand, if $0<\nu\left(e\right)<1$
we say that the environment $\nu$ is \emph{noisy} or \emph{stochastic}.
If we restrict ourselves to deterministic environments such that $\nu\left(e\right)\in\left\{ 0,1\right\} $
$\forall\nu\,\forall e$, then $\xi\left(\cdot\right)\in\left(0,1\right)$
implies that $w_{\nu}\in\left(0,1\right)$ for at least one $\nu\in\mathcal{M}$.
This motivates us to define two agents that seek out percepts to which
the mixture $\xi$ assigns low probability; in deterministic environments,
these will behave like knowledge-seekers. 
\begin{defn}[Square \& Shannon KSA; Orseau, 2011]
\label{def:squareksa}The\emph{ Square} and \emph{Shannon KSA }are
the Bayesian agents with utility $u\left(\ae_{1;t}\right)$ given
by $-\xi\left(\ae_{1:t}\right)$ and $-\log\xi\left(\ae_{1:t}\right)$
respectively. 
\end{defn}
Square, Shannon, and KL-KSA are so-named because of the form of the
expression when one computes the $\xi$-expected utility: this is
clear for Square and Shannon, and for KL it turns out that the expected
information gain is equal to the posterior weighted KL-divergence
$\mathbb{E}_{\xi}\left[\textsc{IG}\right]=\sum_{\nu\in\mathcal{M}}w_{\nu\lvert e}\text{KL}\left(\nu\|\xi\right)$
\cite{Lattimore:2013}. Note that as far as implementation is concerned,
these knowledge-seeking agents differ from AI$\xi$ \emph{only} in
their utility functions. 

The following two algorithms (BayesExp and Thompson sampling) are
RL agents that add exploration heuristics so as to obtain weak asymptotic
optimality, at the cost of introducing an \emph{exploration schedule
}$\left\{ \varepsilon_{1},\varepsilon_{2},\dots\right\} $ which can
be annealed, i.e. $\varepsilon_{t}\leq\varepsilon_{t-1}$ and $\varepsilon_{t}\to0$
as $t\to\infty$. 

\textbf{BayesExp}. BayesExp (Algorithm \ref{alg:bayesexp}) augments
the Bayes agent AI$\xi$ with bursts of exploration, using the information-seeking
policy of KL-KSA. If the expected information gain exceeds a threshold
$\varepsilon_{t}$, the agent will embark on an information-seeking
policy $\pi_{\xi}^{\star,\text{IG}}=\arg\max_{\pi}V_{\xi}^{\pi,\text{IG}}$
for one effective horizon, where $\text{IG}$ is defined in Equation
\eqref{eq:utility-kl}.

\begin{algorithm}
\begin{algorithmic}[1]
\Require{Model class $\mathcal{M}$; prior $w\ :\ \mathcal{M}\to(0,1)$; exploration schedule $\lbrace\varepsilon_1,\varepsilon_2,\dots\rbrace$.
}
\Statex
\Loop
\If{$V_{\xi}^{*,\text{IG}}\left(\ae_{<t}\right)>\varepsilon_t$}
\For{$i = 1\to H_\gamma^t\left(\varepsilon_t\right)$}
\State $\textsc{act}\left(\pi_{\xi}^{\star,\text{IG}}\right)$
\EndFor
\Else
\State $\textsc{act}\left(\pi_{\xi}^{\star}\right)$
\EndIf
\EndLoop
\end{algorithmic}

\caption{BayesExp {[}Lattimore, 2013{]}}
\label{alg:bayesexp}
\end{algorithm}

\textbf{Thompson sampling}. Thompson sampling (TS) is a very common
Bayesian sampling technique, named for \cite{Thompson:1933}. In the
context of general reinforcement learning, it can be used as another
attempt at solving the exploration problems of AI$\xi$. From Algorithm
\ref{alg:thompson}, we see that TS follows the $\rho$-optimal policy
for an effective horizon before re-sampling from the posterior $\rho\sim w\left(\cdot\lvert\ae_{<t}\right)$.
This commits the agent to a single hypothesis for a significant amount
of time, as it samples and tests each hypothesis one at a time. 

\begin{algorithm}[h]
\begin{algorithmic}[1]
\Require{Model class $\mathcal{M}$; prior $w\ :\ \mathcal{M}\to(0,1)$; exploration schedule $\lbrace\varepsilon_1,\varepsilon_2,\dots\rbrace$.
}
\Statex
\Loop
\State Sample $\rho\sim w\left(\cdot\lvert \ae_{<t}\right)$
\State $d\leftarrow H_t\left(\epsilon_t\right)$
\For{$i = 1\to H_t\left(\varepsilon_t\right)$}
\State $\textsc{act}\left(\pi_{\rho}^{\star}\right)$
\EndFor
\EndLoop
\end{algorithmic}

\caption{Thompson Sampling {[}Leike\emph{ et al.}, 2016{]}}
\label{alg:thompson}
\end{algorithm}

\textbf{MDL}. The minimum description length (MDL) principle is an
inductive bias that implements Occam's razor, originally attributed
to \cite{Rissanen:1978}. The MDL agent greedily picks the simplest
probable unfalsified environment $\sigma$ in its model class and
behaves optimally with respect to that environment until it is falsified.
If $\mu\in\mathcal{M}$, Algorithm \ref{alg:mdl} converges with $\sigma\to\mu$.
Note that line $2$ of Algorithm \ref{alg:mdl} amounts to maximum
likelihood regularized by the Kolmogorov complexity $K$.

\begin{algorithm}
\begin{algorithmic}[1]
\Require{Model class $\mathcal{M}$; prior $w\ :\ \mathcal{M}\to(0,1)$, regularizer constant $\lambda\in\mathbb{R}^+$.
}
\Statex
\Loop
\State $\sigma\leftarrow\arg\min_{\nu\in\mathcal{M}}\left[K(\nu) - \lambda\sum_{k=1}^{t}\log\nu(e_k\lvert\ae_{<k}a_k)\right]$
\State $\textsc{act}\left(\pi_{\sigma}^{\star}\right)$
\EndLoop
\end{algorithmic}

\caption{MDL Agent {[}Lattimore and Hutter, 2011{]}. }
\label{alg:mdl}
\end{algorithm}

\section{Implementation}

In this section, we describe the environments that we use in our experiments,
introduce two Bayesian environment models, and discuss some necessary
approximations.

\subsection{Gridworld}

We run our experiments on a class of partially-observable gridworlds.
The gridworld is an $N\times N$ grid of tiles; tiles can be either
empty, walls, or stochastic reward dispenser tiles. The action space
is given by $\mathcal{A}=\left\{ \leftarrow,\rightarrow,\uparrow,\downarrow,\emptyset\right\} $,
which move the agent in the four cardinal directions or stand still.
The observation space is $\mathcal{O}=\mathbb{B}^{4}$, the set of
bit-strings of length four; each bit is $1$ if the adjacent tile
in the corresponding direction is a wall, and is $0$ otherwise. The
reward space is small, and integer-valued: the agent receives $r=-1$
for walking over empty tiles, $r=-10$ for bumping into a wall, and
$r=100$ with some fixed probability $\theta$ if it is on a reward
dispenser tile. There is no observation to distinguish empty tiles
from dispensers. In this environment, the optimal policy (assuming
unbounded episode length) is to move to the dispenser with highest
payout probability $\theta$ and remain there, subsequently collecting
$100\theta$ reward per cycle in expectation (the environment is non-episodic).
In all cases, the agent's starting position is at the top left corner
at tile $\left(0,0\right)$.

This environment has small action and percept spaces and relatively
straightforward dynamics. The challenge lies in coping with perceptual
aliasing due to partial observability, dealing with stochasticity
in the rewards, and balancing exploration and exploitation. In particular,
for a gridworld with $n$ dispensers with $\theta_{1}>\theta_{2}>\theta_{n}$,
the gridworld presents a non-trivial exploration/exploitation dilemma;
see Figure \ref{lab:gridworldz}. 

\subsection{Models}

Generically, we would like to construct a Bayes mixture over a model
class $\mathcal{M}$ that is as rich as possible. Since we are using
a nonparametric model, we are not concerned with choosing our model
class so as to arrange for conjugate prior/likelihood pairs. One might
consider constructing $\mathcal{M}$ by enumerating all $N\times N$
gridworlds of the class described above, but this is infeasible as
the size of such an enumeration explodes combinatorially: using just
two tile types we would run out of memory even on a modest $6\times6$
gridworld since $\left|\mathcal{M}\right|=2^{36}\approx7\times10^{10}$.
Instead, we choose a discrete parametrization $D$ that enumerates
an interesting subset of these gridworlds. One can think of $D$ as
describing a set of parameters about which the agent is uncertain;
all other parameters are held constant, and the agent is fully informed
of their value. We use this to create the first of our model classes,
$\mathcal{M}_{\text{loc}}$. The second, $\mathcal{M}_{\text{Dirichlet}}$,
uses a factorized distribution rather than an explicit mixture to
avoid this issue.

$\boldsymbol{\mathcal{M}_{\text{loc}}}$. This is a Bayesian mixture
parametrized by goal location; the agent knows the layout of the gridworld
and knows its dynamics, but is uncertain about the location of the
dispensers, and must explore the world to figure out where they are.
We construct the model class by enumerating each of these gridworlds.
For square gridworlds of side length $N$, $\left|\mathcal{M}_{\text{loc}}\right|=N^{2}$.
From Algorithm \ref{alg:bayes-rl} the time complexity of our Bayesian
updates is $\mathcal{O}\left(\left|\mathcal{M}\right|\right)=\mathcal{O}\left(N^{2}\right)$.

$\boldsymbol{\mathcal{M}_{\text{Dirichlet}}}$. The Bayes mixture
$\mathcal{M}_{\text{loc}}$ is a natural class of gridworlds to consider,
but it is quite constrained in that it holds the maze layout and dispenser
probabilities fixed. We seek a model that allows the agent to be uncertain
about these features. To do this we move away from the mixture formalism
so as to construct a bespoke gridworld model.

Let $s_{ij}\in\left\{ \textsc{Empty},\textsc{Wall},\textsc{Dispenser},\dots\right\} $
be the state of tile $\left(i,j\right)$ in the gridworld. The joint
distribution over all gridworlds $s_{11},\dots,s_{NN}$ factorizes
across tiles, and the state-conditional percept distributions are
Dirichlet over the four tile types. We effectively use a Haldane prior
\textendash{} $\text{Beta}\left(0,0\right)$ \textendash{} with respect
to $\textsc{Wall}$s and a Laplace prior \textendash{} $\text{Beta}\left(1,1\right)$
\textendash{} with respect to $\textsc{Dispenser}$s. This model class
allows us to make the agent uncertain about the maze layout, including
the number, location, and payout probabilities of $\textsc{Dispenser}$s.
In contrast to $\mathcal{M}_{\text{loc}}$, model updates are $\mathcal{O}(1)$
time complexity, making $\mathcal{M}_{\text{Dirichlet}}$ a far more
efficient choice for large $N$.

This model class incorporates more domain knowledge, and allows the
agent to flexibly and cheaply represent a much larger class of gridworlds
than using the naive enumeration $\mathcal{M}_{\text{loc}}$. Importantly,
$\mathcal{M}_{\text{Dirichlet}}$ is still a Bayesian model \textendash{}
we simply lose the capacity to represent it explicitly as a mixture
of the form of Equation \eqref{eq:mixture}.

\subsection{Agent Approximations}

\textbf{Planning}. In contrast to model-free agents, in which the
policy is explicitly represented by e.g.\ a neural network, our model-based
agents need to calculate their policy from scratch at each time step
by computing the value function in Equation \eqref{eq:bayes-expetimax}.
We approximate this infinite forsight with a finite horizon $m>0$,
and we approximate the expectation using Monte Carlo techniques. We
implement the $\rho$UCT algorithm \cite{SilverVeness2010,VNHUS:2011},
a history-based version of the UCT algorithm \cite{KS2006}. UCT is
itself a variant of Monte Carlo Tree Search (MCTS), a commonly used
planning algorithm originally developed for computer Go. The key idea
is to try to avoid wasting time considering unpromising sequences
of actions; for this reason the action selection procedure within
the search tree is motivated by the upper confidence bound (UCB) algorithm
\cite{AuerCF02}: 
\begin{equation}
a_{\text{UCT}}=\arg\max_{a\in\mathcal{A}}\left(\frac{1}{m\left(\beta-\alpha\right)}\hat{V}\left(\ae_{<t}a\right)+C\sqrt{\frac{\log T\left(\ae_{<t}\right)}{T\left(\ae_{<t}a\right)}}\right),\label{eq:uct}
\end{equation}

where $T\left(\ae_{<t}\right)$ is the number of times the sampler
has reached history $\ae_{<t}$, $\hat{V}\left(\ae_{<t}a\right)$
is the current estimator of $V\left(\ae_{<t}a\right)$, $m$ is the
planning horizon, $C$ is a free parameter (usually set to $\sqrt{2}$),
and $\alpha$ and $\beta$ are the minimum and maximum rewards emitted
by the environment. In this way, the MCTS planner approximates the
expectimax expression in Equation \eqref{eq:expectimax}, and effectively
yields a stochastic policy approximating $\pi$ defined in Equation
\eqref{eq:bayes-expetimax} \cite{VNHUS:2011}. 

In practice, MCTS is very computationally expensive; when planning
by forward simulation with a Bayes mixture $\xi$ over $\mathcal{M}$,
the worst-case time-complexity of $\rho$UCT is $\mathcal{O}\left(\kappa m\left|\mathcal{M}\right|\left|\mathcal{A}\right|\right)$,
where $\kappa$ is the Monte Carlo sample budget. It is important
to note that $\rho$UCT treats the environment model $\xi$ as a black
box: it is agnostic to the environment's structure, and so makes no
assumptions or optimizations. For this reason, planning in POMDPs
with $\rho$UCT is quite wasteful: due to perceptual aliasing, the
algorithm considers many plans that are cyclic in the (hidden) state
space of the POMDP. This is unavoidable, and means that in practice
$\rho$UCT can be very inefficient even in small environments. 

\textbf{Effective horizon}.\textbf{ }Computing the effective horizon
exactly for general discount functions is not possible in general,
although approximate effective horizons have been derived for some
common choices of $\gamma$ \cite{Lattimore:2013}. For most realistic
choices of $\gamma$ and $\varepsilon$, the effective horizon $H_{\gamma}\left(\varepsilon\right)$
is significantly greater than the planning horizon $m$ we can feasibly
use due to the prohobitive computational requirements of MCTS \cite{LALH:2017}.
For this reason we use the MCTS planning horizon $m$ as a surrogate
for $H_{\gamma}^{t}$. In practice, all but small values of $m$ are
feasible, resulting in short-sighted agents with (depending on the
environment and model) suboptimal and highly stochastic policies. 

\textbf{Utility bounds}. Recall from Equation \eqref{eq:uct} that
the $\rho$UCT value estimator $\hat{V}\left(\ae_{1:t}\right)$ is
normalized by a factor of $m\left(\beta-\alpha\right)$. For reward-based
reinforcement learners, $\alpha$ and $\beta$ are part of the metadata
provided to the agent by the environment at the beginning of the agent-environment
interaction. For utility-based agents such as KSA, however, rewards
are generated intrinsically, and so the agent must calculate for itself
what range of utilities it expects to see, so as to correctly normalize
its value function for the purposes of planning. For Square and KL-KSA,
it is possible to get loose utility bounds $\left[\alpha,\beta\right]$
by making some reasonable assumptions. Since $u_{\text{Square}}\left(e\right)=-\xi\left(e\right)$,
we have the bound $-1\leq u_{\text{Square}}\leq0$. One can argue
that for the vast majority of environments this bound will be tight
above, since there will exist percepts $e_{f}$ that the agent's model
has effectively falsified such that $\xi\left(e_{f}\right)\to0$ as
$t\to\infty$. 

In the case of the KL-KSA, recall that $u_{\mbox{KL}}\left(e\right)=\text{Ent}\left(w\left(\cdot\right)\right)-\text{Ent}\left(w\left(\cdot\lvert e\right)\right)$.
If we assume a uniform prior $w\left(\cdot\lvert\epsilon\right)$,
then we have $0\leq u_{\mbox{KL}}\left(e\right)\leq\text{Ent}\left(w\left(\cdot\lvert\epsilon\right)\right)\ \forall e\in\mathcal{E}$
since entropy is non-negative over discrete model classes. 

Now, in general $u_{\mbox{Shannon}}=-\log\xi$ is unbounded above
as $\xi\to0$, so unless we can \emph{a priori} place lower bounds
on the probability that $\xi$ will assign to an arbitrary percept
$e\in\mathcal{E}$, we cannot bound its utility function and therefore
cannot calculate the normalization in Equation \eqref{eq:uct}. Therefore,
planning with MCTS with the Shannon KSA will be problematic in many
environments, as we are forced to make an arbitrary choice of upper
bound $\beta$.

\section{Experiments}

We run experiments to investigate and compare the agents' learning
performance. Except where otherwise stated, the following experiments
were run on a $10\times10$ gridworld with a single dispenser with
$\theta=0.75$. We average training score over $50$ simulations for
each agent configuration, and we use $\kappa=600$ MCTS samples and
a planning horizon of $m=6$. In all cases, discounting is geometric
with $\gamma=0.99$. In all cases, the agents are initialized with
a uniform prior.

\begin{figure}
\begin{centering}
\includegraphics[scale=0.25]{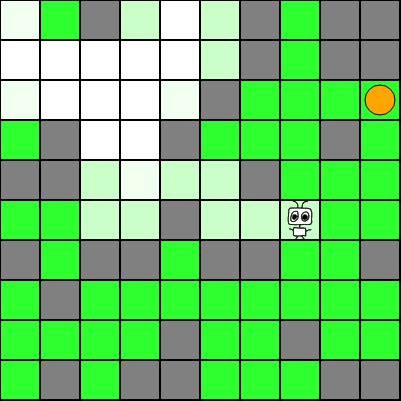}$\qquad$\includegraphics[scale=0.185]{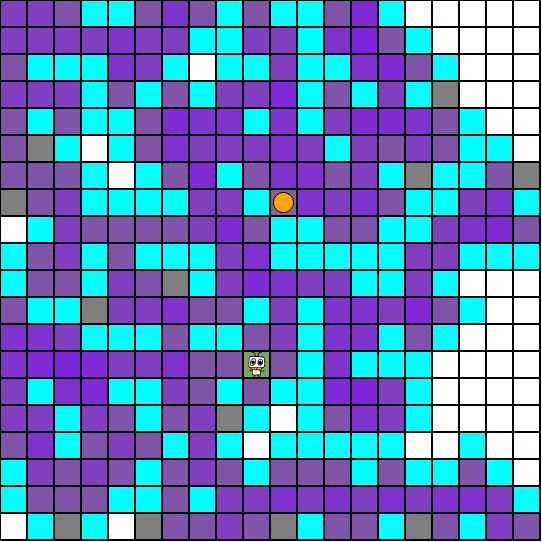}
\par\end{centering}
\caption{\textbf{Left}:\textbf{ }An example $10\times10$ gridworld environment,
with the agent's posterior $w_{\nu\lvert\ae_{<t}}$ over $\mathcal{M}_{\text{loc}}$
superimposed. Grey tiles are walls, and the agent's posterior is visualized
in green; darker shades correspond to tiles with greater probability
mass. Here, AI$\xi$ has mostly falsified the hypotheses that the
goal is in the top-left corner of the maze, and so is motivated to
search deeper in the maze, as its model assigns greater mass to unexplored
areas. In the ground truth environment $\mu$, the dispenser is located
at the tile represented by the orange disc. \textbf{Right}: A $20\times20$
gridworld that is too large to feasibly model with $\mathcal{M}_{\text{loc}}$,
with the agent's posterior over $\mathcal{M}_{\text{Dirichlet}}$
superimposed. White tiles are unknown, pale blue tiles are known to
be walls, and purple tiles are known to not be walls; darker shades
of purple correspond to lower credence that a $\textsc{Dispenser}$
is present. Notice that despite exploring much of the maze, AI$\xi$
has not discovered the profitable reward dispenser located in the
upper centre; it has instead settled for a suboptimal dispenser in
the lower part of the maze, illustrating the exploration-exploitation
tradeoff.}

\centering{}\label{lab:gridworldz}
\end{figure}

We plot the training score averaged over $N=50$ simulation runs,
with shaded areas corresponding to one sigma. That is, we plot $\bar{s}_{t}=\frac{1}{tN}\sum_{i=1}^{N}\sum_{j=1}^{t}s_{ij}$
where $s_{ij}$ is the instantaneous score at time $j$ in run $i$:
this is either reward in the case of reinforcement learners, or fraction
of the environment explored in the case of KSA.

\begin{figure}[t]
\begin{centering}
\includegraphics[scale=0.16]{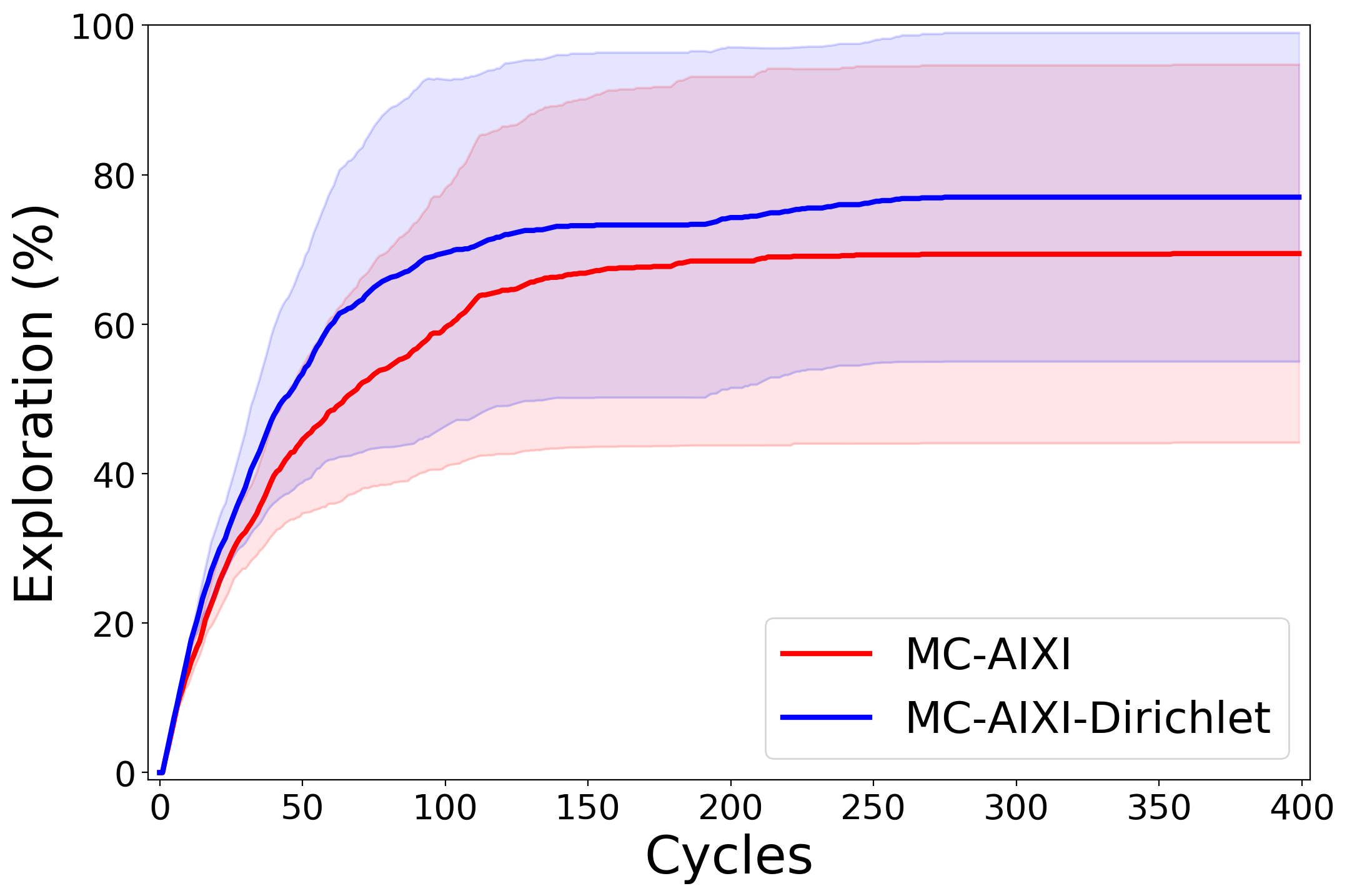}\includegraphics[scale=0.16]{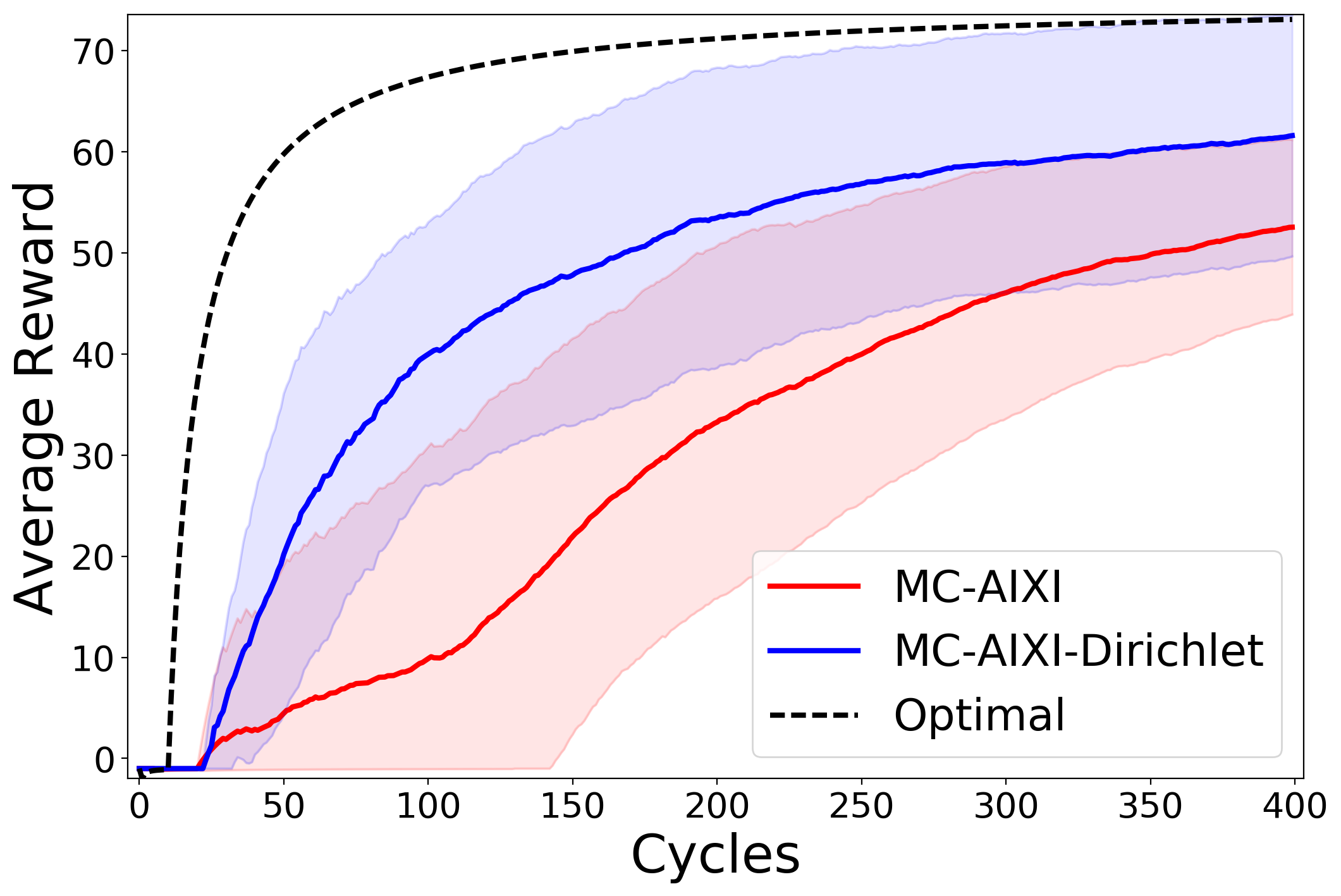}
\par\end{centering}
\caption{Performance of AI$\xi$ is dependent on model class ($\mathcal{M}_{\text{loc}}$
in red, $\mathcal{M}_{\text{Dirichlet}}$ in blue, `Cycles'$\equiv t$).
\textbf{Left}: Exploration fraction, $f=100\times\frac{N_{\text{visited }}}{N_{\text{reachable}}}$.
Note that MC-AIXI and MC-AIXI-Dirichlet both effectively stop exploring
quite early on, at around $t\approx150$. \textbf{Right}: Average
reward. }

\centering{}\label{fig:comparison-aixi-models}
\end{figure}

\textbf{Model class}. AI$\xi$ performs significantly better using
the Dirichlet model than with the mixture model. Since the Dirichlet
model is less constrained (in other words, less informed), there is
more to learn, and the agent is more motivated to explore, and is
less susceptible to getting stuck in local maxima. From Figure \ref{fig:comparison-aixi-models},
we see that MC-AIXI-Dirichlet appears to have asymptotically higher
variance in its average reward than MC-AIXI. This makes sense since
the agent may discover the reward dispenser, but be subsequently incentivized
to move away from it and keep exploring, since its model still assigns
significant probability to there being better dispensers elsewhere;
in contrast, under $\mathcal{M}_{\text{loc}}$, the agent's posterior
immediately collapses to the singleton once it discovers a dispenser,
and it will greedily stay there ever after. This is borne out by Figure
\ref{fig:comparison-aixi-models}, which shows that, on average, AIXI
explores significantly more of the gridworld using $\mathcal{M}_{\text{Dirichlet}}$
than with $\mathcal{M}_{\text{loc}}$. These experiments illustrate
how AI$\xi$'s behavior depends strongly on the model class $\mathcal{M}$.

\textbf{KSA}. As discussed in Section \ref{subsec:algorithms}, the
Shannon- and Square-KSA are entropy-seeking; they are therefore susceptible
to environments in which a noise source is constructed so as to trap
the agent in a very suboptimal exploratory policy, as the agent gets
`hooked on noise' \cite{OLH:2013ksa}. The noise source is a tile
that emits uniformly random percepts over a sufficiently large alphabet
such that the probability of any given percept $\xi\left(e\right)$
is lower (and more attractive) than anything else the agent expects
to experience by exploring. 

\begin{figure}[h]
\begin{centering}
\includegraphics[scale=0.16]{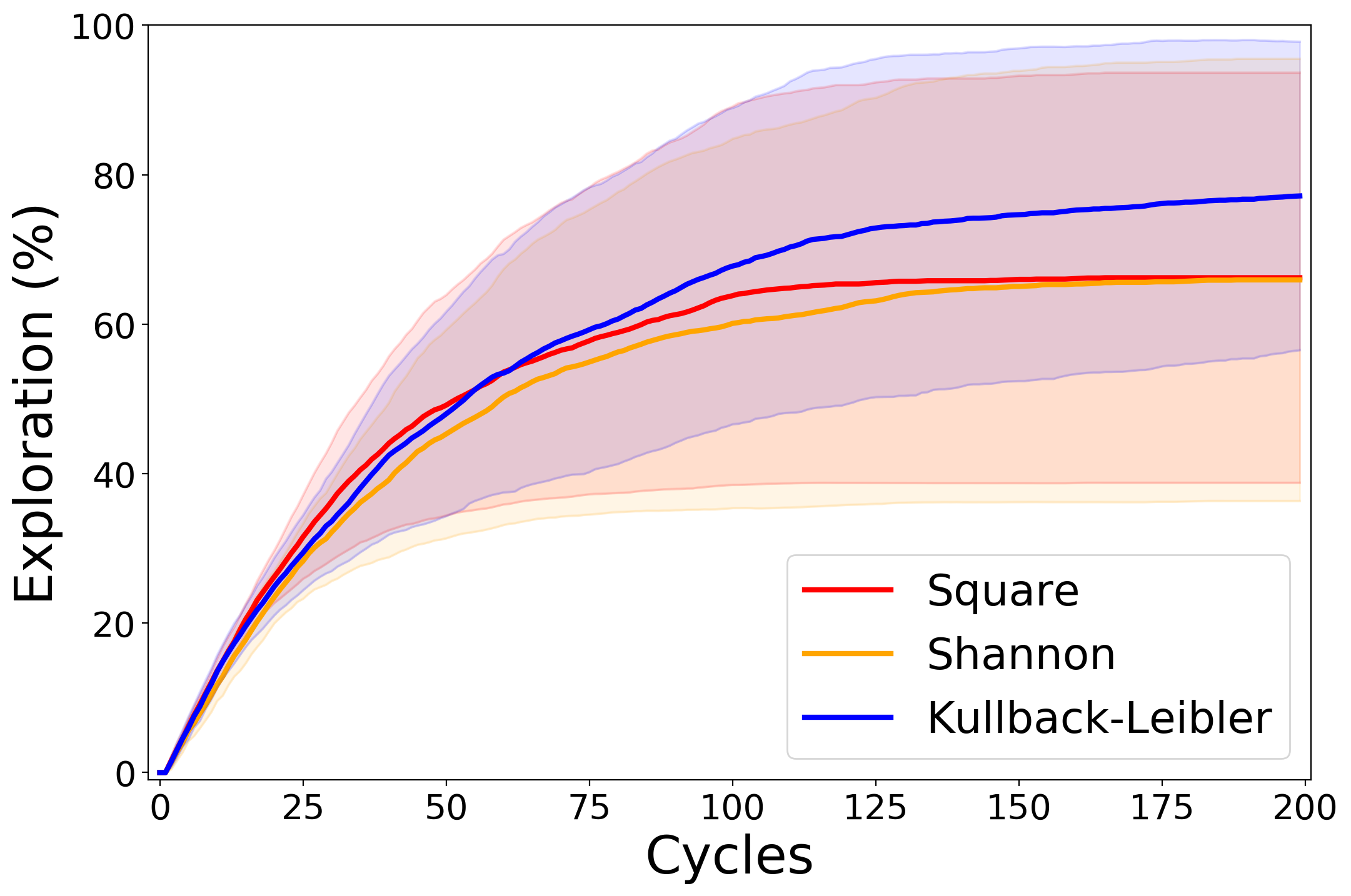}\includegraphics[scale=0.16]{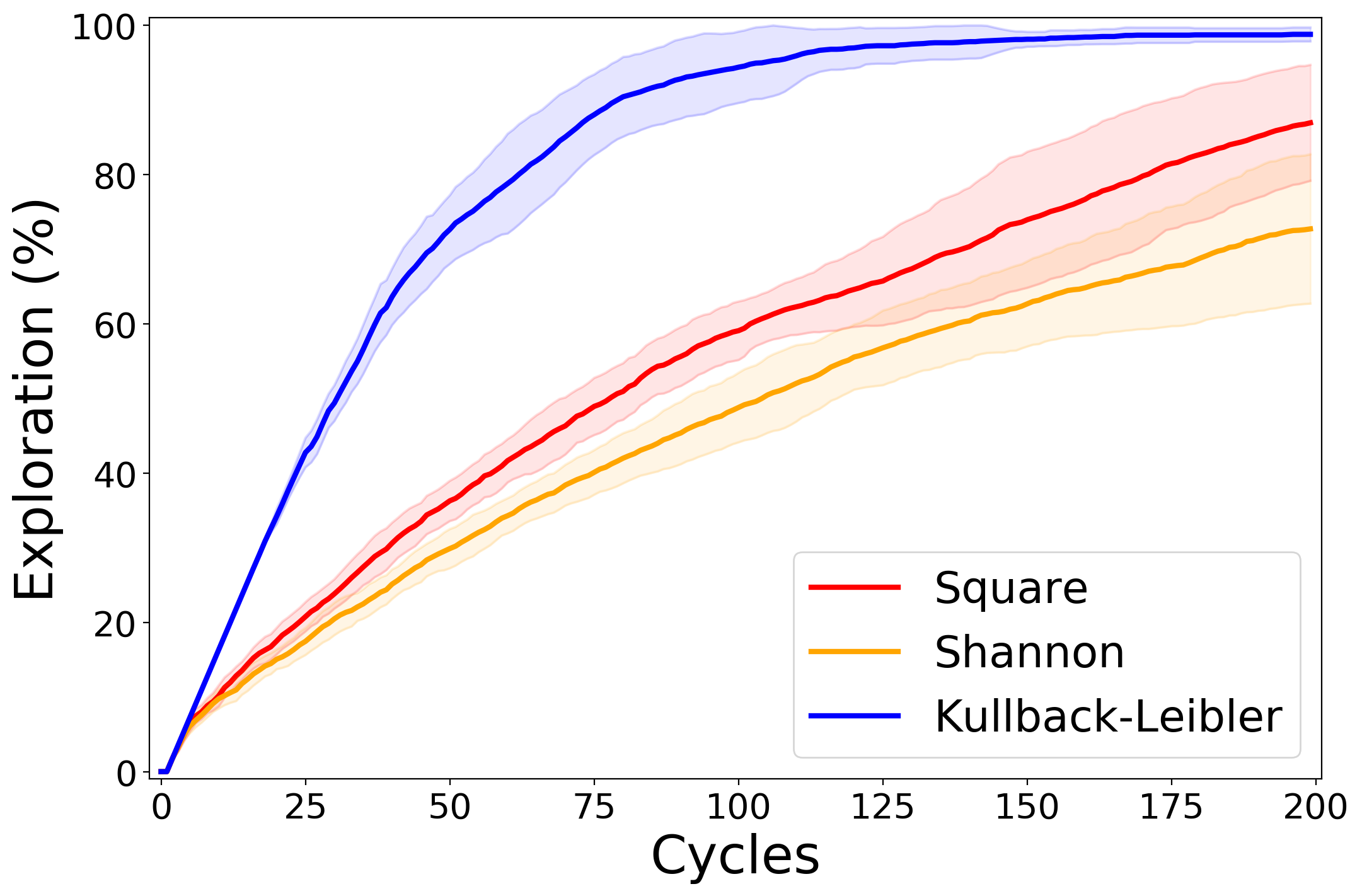}
\par\end{centering}
\caption{Intrinsic motivation is highly model-dependent, and the information-seeking
policy outperforms entropy-seeking in stochastic environments. \textbf{Left}:
$\mathcal{M}_{\text{loc}}$.\textbf{ Right}: $\mathcal{M}_{\text{Dirichlet}}$. }

\centering{}\label{fig:comparison-ksa}
\end{figure}

\begin{figure}[b]
\begin{centering}
\includegraphics[scale=0.14]{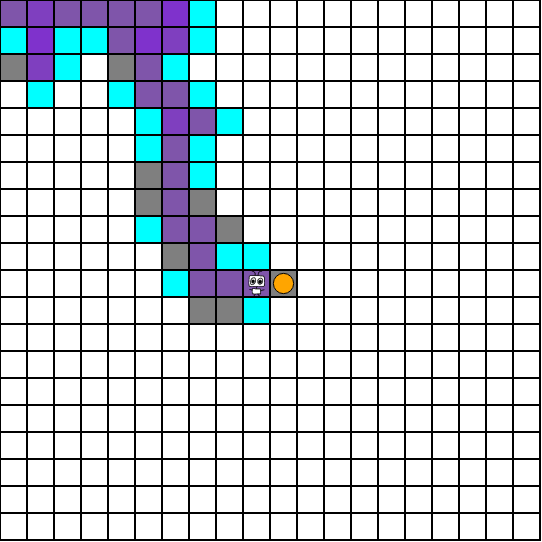}$\ $\includegraphics[scale=0.14]{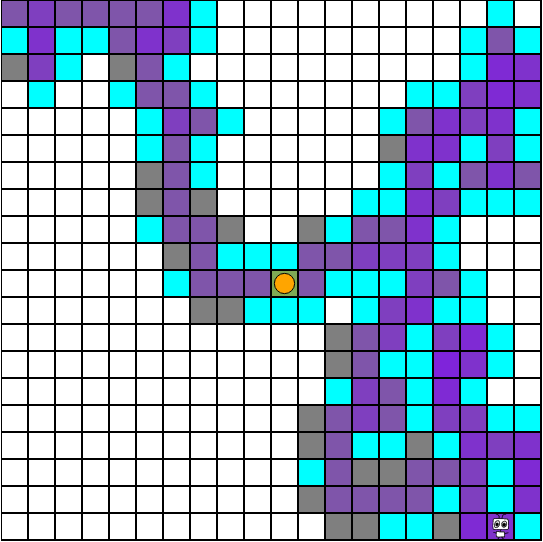}$\ $\includegraphics[scale=0.14]{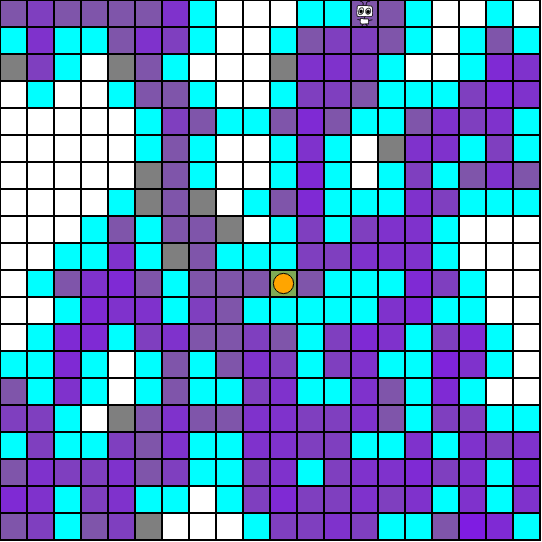}
\par\end{centering}
\caption{KL-KSA-Dirichlet is highly motivated to explore every reachable tile
in the gridworld. \textbf{Left} ($t=36$): The agent begins exploring,
and soon stumbles on the dispenser tile. \textbf{Center }($t=172$):
The agent is motivated only by information gain, and so ignores the
reward dispenser, opting instead to continue exploring the maze. \textbf{Right}
($t=440$): The agent has now discovered $>90\%$ of the maze, and
continues to gain information from tiles it has already visited as
it updates its independent Laplace estimators for each tile.}

\label{fig:ksa-explores}
\end{figure}

KL-KSA explores more effectively than Square- and Shannon-KSA in stochastic
environments; see Figure \ref{fig:comparison-ksa}. Under the mixture
model $\mathcal{M}_{\text{loc}}$, the posterior collapses to a singleton
once the agent finds the goal. Given a stochastic dispenser, this
becomes the only source of entropy in the environment, and so Square-
and Shannon-KSA will remain at the dispenser. In contrast, once the
posterior collapses to the minimum entropy configuration, there is
no more information to be gained, and so KL-KSA will random walk (breaking
ties randomly), and outperform Square and Shannon-KSA marginally in
exploration. This difference becomes more marked under the Dirichlet
model; although all three agents perform better under $\mathcal{M}_{\text{Dirichlet}}$
due to its being less constrained and having more entropy than $\mathcal{M}_{\text{loc}}$,
KL-KSA outperforms the others by a considerable margin; KL-KSA is
motivated to explore as it anticipates considerable changes to its
model by discovering new areas; see Figure \ref{fig:ksa-explores}.

\begin{figure}[h]
\begin{centering}
\includegraphics[scale=0.16]{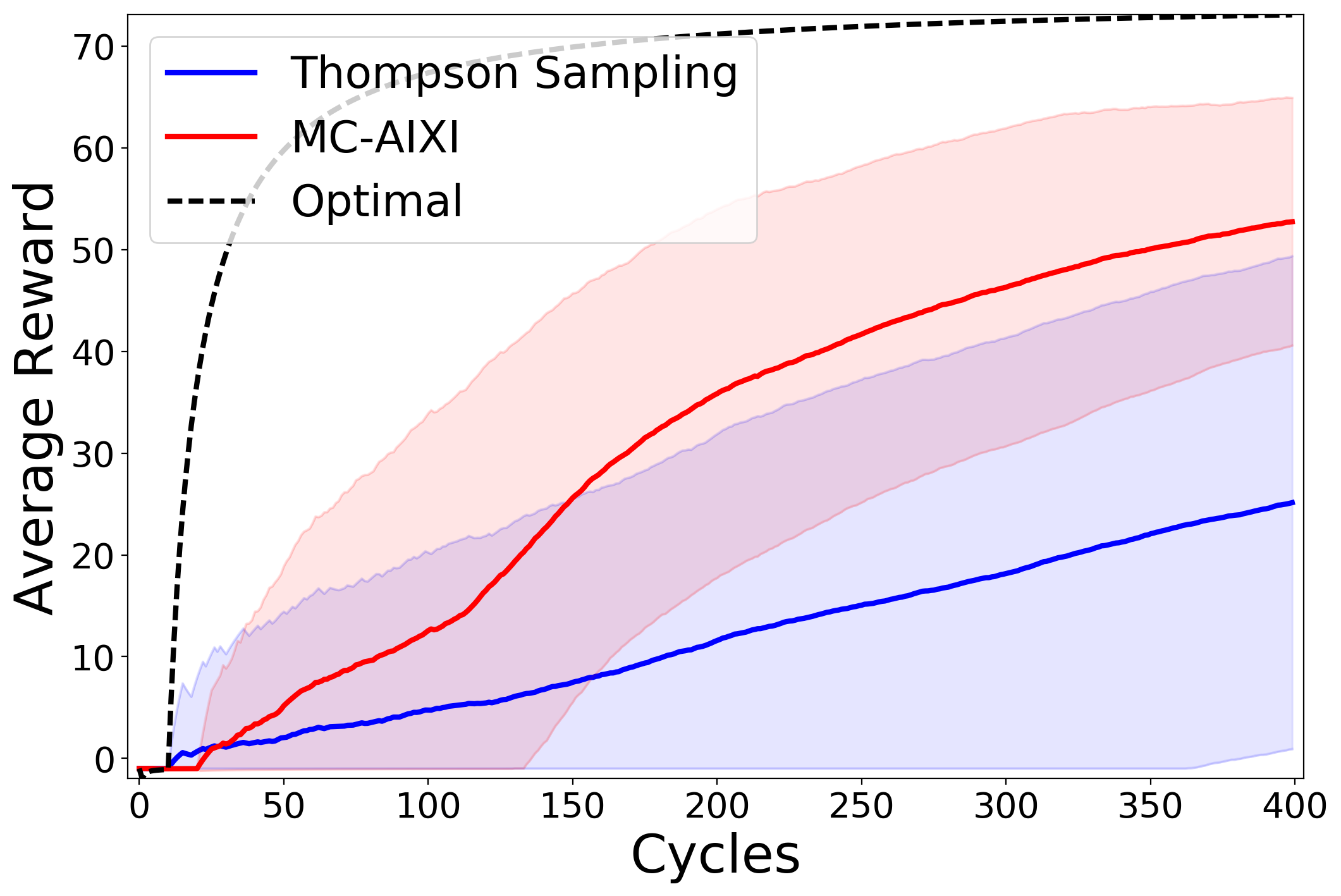}\includegraphics[scale=0.16]{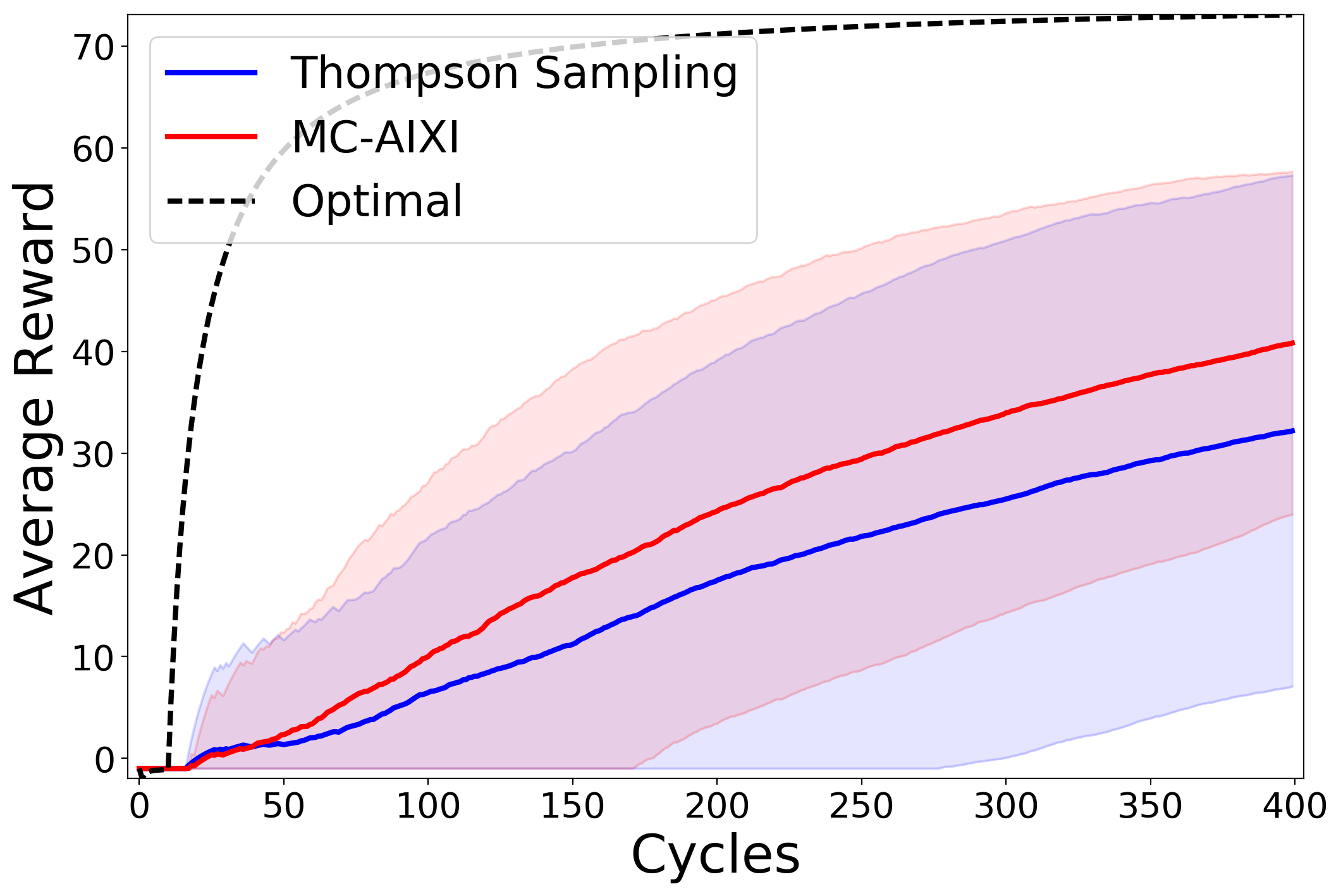}
\par\end{centering}
\caption{TS typically underperforms AI$\xi$. \textbf{Left}: Both agents are
given the same MCTS planning parameters: $m=6$, and a samples budget
of $\kappa=600$. Here, Thompson sampling is unreliable and lacklustre,
achieving low mean reward with high variance.\textbf{ Right}:\textbf{
}When each agent is given a horizon of $m=10$ and a time budget of
$t_{\text{max}}=100$ milliseconds per action, TS performs comparatively
better, as the gap to AI$\xi$ is narrowed.}

\centering{}\label{fig:comparison-thompson}
\end{figure}

\textbf{Exploration}. Thompson sampling (TS) is asymptotically optimal
in general environments, a property that AI$\xi$ lacks. However,
in our experiments on gridworlds, TS performs poorly in comparison
to AI$\xi$; see Figure \ref{fig:comparison-thompson}. This is caused
by two issues: (1) The parametrization of $\mathcal{M}_{\text{loc}}$
means that the $\rho$-optimal policy for any $\rho\in\mathcal{M}_{\text{loc}}$
is to seek out some tile $\left(i,j\right)$ and wait there for one
planning horizon $m$. For all but very low values of $\theta$ or
$m$, this is an inefficient strategy for discovering the location
of the dispenser. (2) The performance of TS is strongly constrained
by limitations of the planner. If the agent samples an environment
$\rho$ which places the dispenser outside its planning horizon \textendash{}
that is, more than $m$ steps away \textendash{} then the agent will
not be sufficiently far-sighted to do anything useful. In contrast,
AI$\xi$ is motivated to continually move around the maze as long
as it hasn't yet discovered the dispenser, since $\xi$ will assign
progressively higher mass to non-visited tiles as the agent explores
more.

Thompson sampling is computationally cheaper than AI$\xi$ due to
the fact that it plans with only one environment model $\rho\in\mathcal{M}$,
rather than with the entire mixture $\xi$. When given a level-playing
field with a time budget rather than a samples budget, the performance
gap is reduced; see Figure \ref{fig:comparison-thompson}b.

\begin{figure}[b]
\begin{centering}
\includegraphics[scale=0.16]{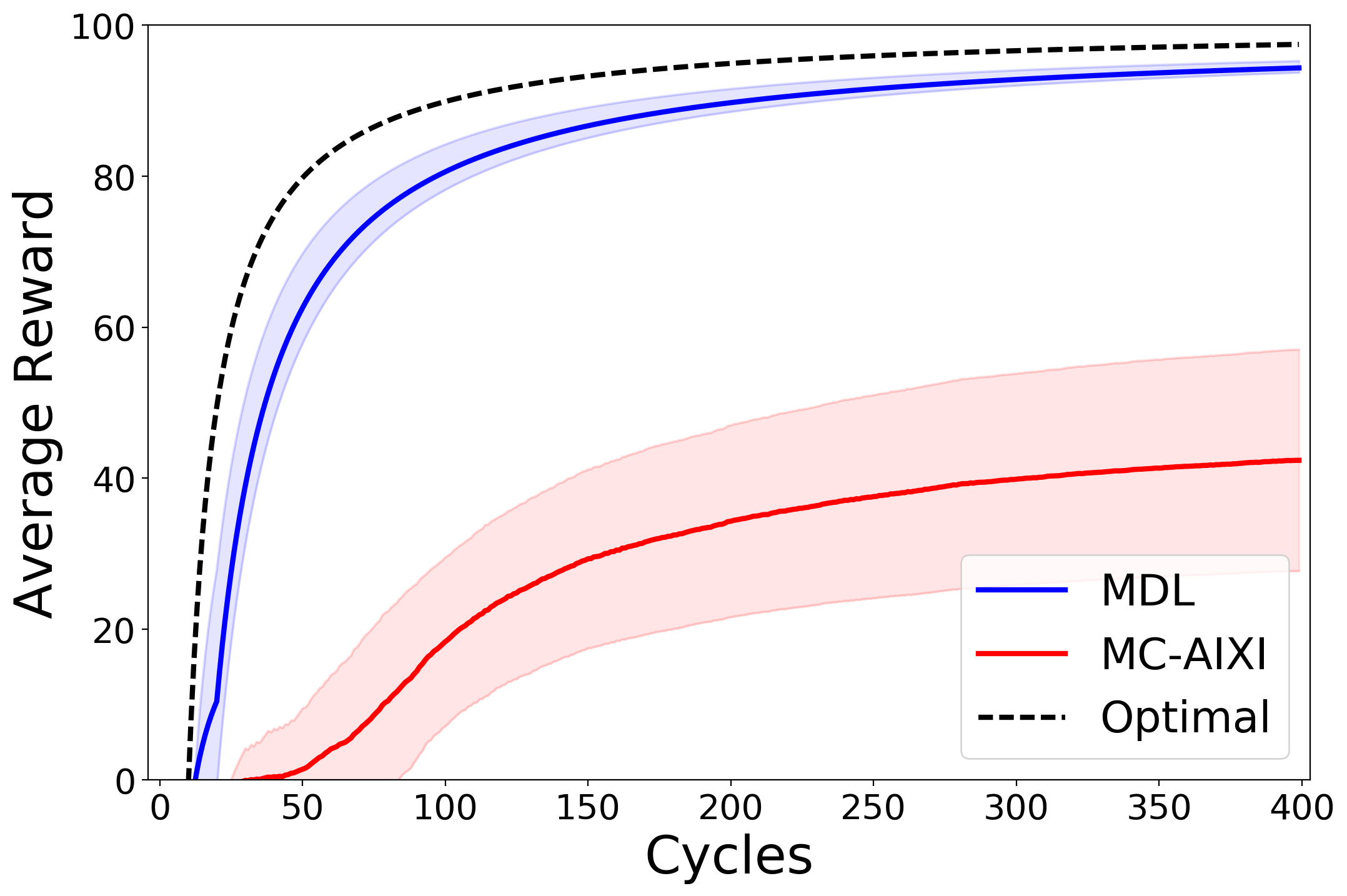}\includegraphics[scale=0.16]{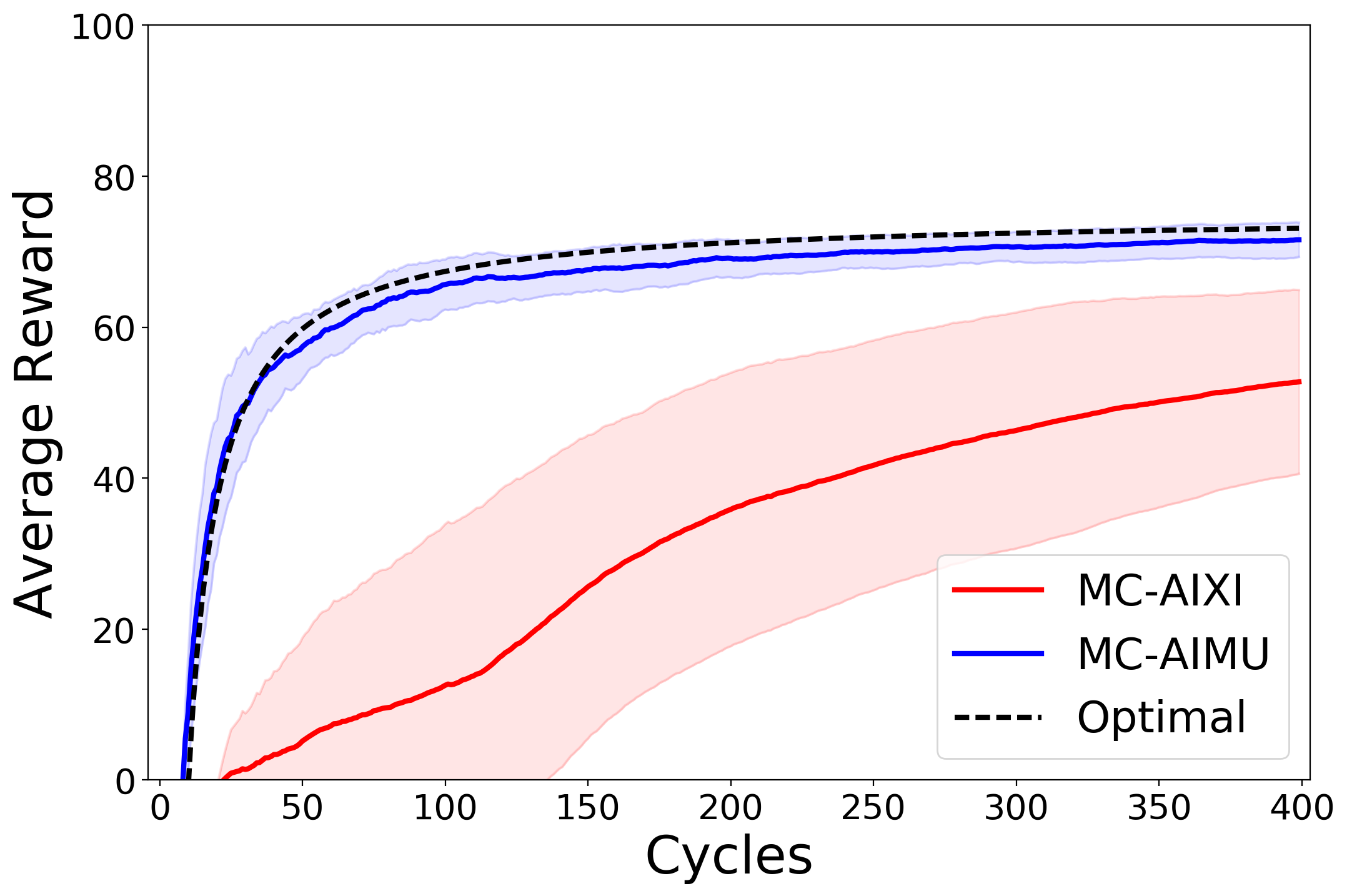}
\par\end{centering}
\caption{\textbf{Left}: MDL vs uniform-prior AI$\xi$ in a deterministic ($\theta=1$)
gridworld. \textbf{Right}: AI$\xi$ compared to the (MC-approximated)
optimal policy AI$\mu$ with $\theta=0.75$. Note that MC-AIXI paradoxically
performs worse in the deterministic setting than the stochastic one;
this is because the posterior over $\xi$ very quickly becomes sparse
as hypotheses are rejected, making planning difficult for small $m$.}

\centering{}\label{fig:comparison-mdl-aixi}
\end{figure}

\textbf{Occam bias}. Since each environment $\nu\in\mathcal{M}_{\text{loc}}$
differs by a single parameter and has otherwise identical source code,
we cannot use a Kolmogorov complexity surrogate to differentiate them.
Instead we opt to order by the environment's index in a row-major
order enumeration. On simple deterministic environments (effectively,
$\lambda=0$ in Algorithm \ref{alg:mdl}), the MDL agent significantly
outperforms the Bayes agent AI$\xi$ with a uniform prior over $\mathcal{M}$;
see Figure \ref{fig:comparison-mdl-aixi}. This is because the MDL
agent is biased towards environments with low indices; using the $\mathcal{M}_{\mbox{loc}}$
model class, this corresponds to environments in which the dispenser
is close to the agent's starting position. In comparison, AI$\xi$'s
uniform prior assigns significant probability mass to the dispenser
being deep in the maze. This motivates it to explore deeper in the
maze, often neglecting to thoroughly exhaust the simpler hypotheses.

\section{Conclusion}

In this paper we have presented a short survey of state-of-the-art
universal reinforcement learning algorithms, and developed experiments
that demonstrate the properties of their resulting exploration strategies.
We also discuss the dependence of the behavior of Bayesian URL agents
on the construction of the model class $\mathcal{M}$, and describe
some of the tricks and approximations that are necessary to make Bayesian
agents work with $\rho$UCT planning. We have made our implementation
open source and available for further development and experimentation,
and anticipate that it will be of use to the RL community in the future.

\section*{Acknowledgements}

We wish to thank Sean Lamont for his assistance in developing the
gridworld visualizations used in Figures \ref{lab:gridworldz} and
\ref{fig:ksa-explores}. We also thank Jarryd Martin and Suraj Narayanan
S for proof-reading early drafts of the manuscript. This work was
supported in part by ARC DP150104590.

\pagebreak{}

\appendix
\bibliographystyle{style/named}
\bibliography{bib/ai}

\end{document}